\documentclass[runningheads]{llncs}
\usepackage{graphicx}

\usepackage{appendix}
\usepackage{tikz}
\usepackage{comment}
\usepackage{amsmath,amssymb} % define this before the line numbering.
\usepackage{color}
\usepackage{mmstyle}
\usepackage{microtype}
\usepackage[rightcaption]{sidecap}
\sidecaptionvpos{figure}{c}
\usepackage{tabulary,multirow,xspace,xcolor}
\usepackage{fixmath,mathtools,nicefrac}

\usepackage[breaklinks=true,colorlinks,bookmarks=false]{hyperref}

\begin{document}
% \linenumbers
\pagestyle{headings}
\mainmatter
\def\ECCVSubNumber{2272}  % Insert your submission number here

\title{Side-Aware Boundary Localization for \\ More Precise Object Detection} % Replace with your title

\begin{comment}
\titlerunning{ECCV-20 submission ID \ECCVSubNumber}
\authorrunning{ECCV-20 submission ID \ECCVSubNumber}
\author{Anonymous ECCV submission}
\institute{Paper ID \ECCVSubNumber}
\end{comment}
%******************

%\begin{comment}
\titlerunning{Side-Aware Boundary Localization for More Precise Object Detection}
\author{Jiaqi Wang\inst{1} \and
	Wenwei Zhang\inst{2} \and
	Yuhang Cao\inst{1} \and
	Kai Chen\inst{3} \and
	Jiangmiao Pang\inst{4} \and \\
	Tao Gong\inst{5} \and
	Jianping Shi\inst{3} \and
	Chen Change Loy \index{Loy, Chen Change}\inst{2} \and
	Dahua Lin\inst{1} \\
}
\authorrunning{J. Wang et al.}
\institute{$^1$ The Chinese University of Hong Kong \\
$^2$ Nanyang Technological University \quad
$^3$ SenseTime Research \\ $^4$ Zhejiang University \quad $^5$ University of Science and Technology of China \\
\email{\tt\small \{wj017,dhlin\}@ie.cuhk.edu.hk}
{\tt\small \{yhcao6,chenkaidev,pangjiangmiao,gongtao950513\}@gmail.com} \\
{\tt\small \{wenwei001,ccloy\}@ntu.edu.sg}
{\quad \tt\small shijianping@sensetime.com}}
%\end{comment}
%******************
\maketitle

%%%%%%%%%%%%%%%%%%%%%%%%%%%%%%%%%%%%%%%%%%%%%%%%%%%%%%%%%%%%%%%%%%%%%%%%%%%%%%%%%%%%%%%%%%%%%%%%%%%%
% !TEX root = ../2272.tex
\begin{abstract}

Current object detection frameworks mainly rely on bounding box regression to localize objects. 
Despite the remarkable progress in recent years, the precision 
of bounding box regression remains unsatisfactory, hence limiting performance in object detection.
We observe that precise localization requires careful placement of each side of 
the bounding box. However, the mainstream approach, which focuses on predicting centers and 
sizes, is not the most effective way to accomplish this task, especially when there exists displacements with large variance between the anchors and the targets.
In this paper, we propose an alternative approach, named as \emph{\textbf{Side-Aware Boundary Localization (SABL)}}, where each side of the bounding box 
is respectively localized with a dedicated network branch. 
To tackle the difficulty of precise localization in the presence of displacements with large variance, 
we further propose a two-step localization scheme, which first predicts a range of movement through 
bucket prediction and then pinpoints the precise position within the predicted bucket. 
We test the proposed method on both two-stage and single-stage detection frameworks. 
Replacing the standard bounding box regression branch with the proposed design leads to 
significant improvements on Faster R-CNN, RetinaNet, and Cascade R-CNN, 
by 3.0\%, 1.7\%, and 0.9\%, respectively. Code is available at \url{https://github.com/open-mmlab/mmdetection}.

\end{abstract}

% !TEX root = ../2272.tex
\section{Introduction}

The development of new frameworks for object detection, 
\eg,~\emph{Faster R-CNN}~\cite{ren2015faster}, \emph{RetinaNet}~\cite{lin2017_focal}, and 
\emph{Cascade R-CNN}~\cite{cascade_rcnn}, has substantially pushed forward 
the state of the art. 
All these frameworks, despite the differences in their technical designs, 
have a common component, namely \emph{bounding box regression}, for object localization.

Generally, bounding box regression is trained to align nearby proposals
to target objects. In a common design, the bounding box regression branch predicts the offsets of the centers $(\delta x, \delta y)$ together 
with the relative scaling factors $(\delta w, \delta h)$ based on the 
features of RoI (Region of Interest).
While this design has been shown to be quite effective in previous works,
it remains very difficult to \emph{precisely} predict
the location of an object when there exists a displacement, with large variance, between 
the anchor and the target. This difficulty also limits the overall detection performance.  

\begin{figure}[t]
	\centering
	\includegraphics[width=0.65\linewidth]{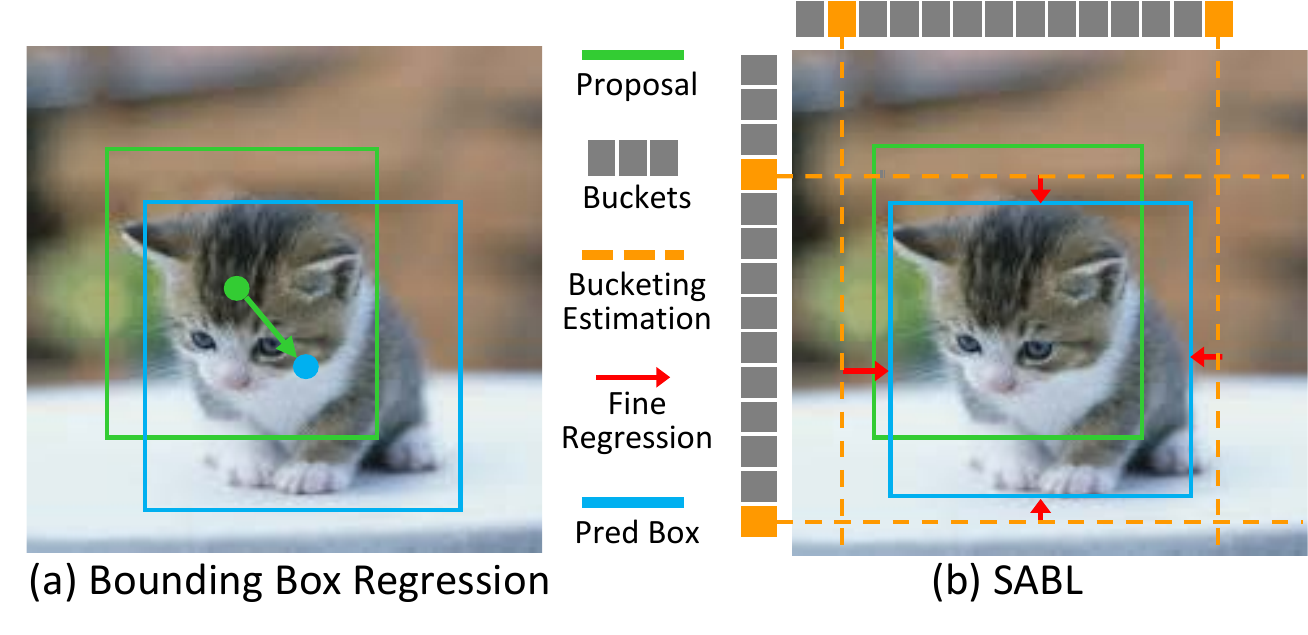}
	\caption{\small{\textbf{The Illustration of Side-Aware Boundary Localization (SABL)}. (a) Common \emph{Bounding box Regression} directly predicts displacements from proposals to ground-truth boxes. (b) SABL focuses on object boundaries and localizes them with a bucketing scheme comprising two steps: bucketing estimation and fine regression }
	}
	\label{fig:motivation}
\end{figure}

In recent years, various efforts have been devoted to improving the localization precision, 
such as
cascading the localization process~\cite{cascade_rcnn,Gidaris_2016,jiang2018acquisition,wang2019region}, and
treating localization as a procedure to segment grid points~\cite{lu2019grid}.
Although being shown effective in boosting the accuracy of localization, 
adoption of these methods complicates the detection pipeline, resulting in considerable 
computational overhead. 

In this work, 
we aim to explore a new approach to object localization that can  
effectively tackle \emph{precise localization} with a lower overhead.
Empirically, we observe that when we manually annotate a bounding box for an object, it is often much 
easier to align each side of the box to the object boundary than to move
the box as a whole while tuning the size. 
Inspired by this observation, we propose a new design, named as \emph{\textbf{Side-Aware Boundary Localization (SABL)}}, where each side of the bounding box
is respectively positioned based on its surrounding context.
As shown in Figure~\ref{fig:motivation}, we devise a \emph{bucketing} scheme to improve the localization precision. For each side of a bounding box, this scheme divides the target space into multiple \emph{buckets}, then determines the bounding box via two steps. Specifically, it first searches for the \emph{correct} bucket, \ie, the one in which the boundary resides. Leveraging the centerline of the selected buckets as a coarse estimate, fine regression is then performed by predicting the offsets. 
This scheme allows very precise localization even in the presence
of displacements with large variance.
Moreover, to preserve precisely localized bounding boxes in the non-maximal 
suppression procedure, we also propose to adjust the classification score based on 
the bucketing confidences, which leads to further performance gains.

We evaluate the proposed SABL upon various detection frameworks, including 
two-stage~\cite{ren2015faster}, single-stage~\cite{lin2017_focal}, and cascade~\cite{cascade_rcnn} detectors. 
By replacing the existing bounding box regression branch with the proposed design, 
we achieve significant improvements on \emph{COCO test-dev}~\cite{lin2014coco} without inflicting high computational cost, 
\ie~41.8\% vs. 38.8\% $AP$ with only around 10\% extra inference time on top of Faster R-CNN, 
40.5\% vs. 38.8\% $AP$ \emph{without} extra inference time on top of RetinaNet.
Furthermore, we integrate SABL into Cascade R-CNN, where SABL achieves consistent performance gains on this strong baseline, \ie, 43.3\% vs. 42.4\% $AP$.

% !TEX root = ../2272.tex
\section{Related Work}
\noindent\textbf{Object Detection.}
Object detection is one of the fundamental tasks for computer vision applications~\cite{Xiong_2019_ICCV,Huang_2018_CVPR,Wenwei_2019_ICCV,zhan2020self}. Recent years have witnessed a dramatic improvement in object detection~\cite{nas_fpn,Zhao2019_M2Det,yang2019reppoints,Zhu_2019_CVPR,Choi_2019_ICCV,Wang_2019_ICCV,Cao_2020_CVPR,STLattice2018CVPR}.
The two-stage pipeline~\cite{girshick2015fast,ren2015faster} has been the leading paradigm in this area.
The first stage generates a set of region proposals, and then the second stage classifies and refines the coordinates
of proposals by bounding box regression.
This design is widely adopted in the later two-stage methods~\cite{dai2016r,mask_rcnn}.
Compared to two-stage approaches, the single-stage pipeline~\cite{lin2017_focal,liu2016_ssd,Redmon_2016,Redmon_2017} predicts bounding boxes directly.
Despite omission of the proposal generation process, single-stage methods~\cite{lin2017_focal,liu2016_ssd,FreeAnchor} require densely distributed anchors produced by sliding window.
Recently, some works attempt to use anchor-free methods~\cite{kong2019foveabox,Law2018_CornerNet,tian2019fcos} for object detection.
Intuitively, iteratively performing the classification and regression process could effectively improve the detection performance.
Therefore, many attempts~\cite{cascade_rcnn,Gidaris_2016,jiang2018acquisition,wang2019region,najibi2016g,zhang2018single} apply cascade architecture to regress bounding boxes iteratively for progressive refinement.

\noindent\textbf{Object Localization.}
Object localization is one of the crucial and fundamental modules for object detection.
A common approach for object localization is to regress the center coordinate and the size of a bounding box~\cite{dai2016r,girshick2015fast,liu2016_ssd,ren2015faster,girshick2014rich}.
This approach is widely adopted, yet the precision is unsatisfactory due to the large variance of regression target.
Aiming for a more accurate localization, some methods~\cite{cascade_rcnn,jiang2018acquisition,wang2019region,najibi2016g,zhang2018single,HTC} directly repeat the bounding box regression multiple times to further improve accuracy. 
However, such cascading pipeline expenses much more computational overhead.
Some methods that try to reformat the object localization process.
Grid R-CNN~\cite{lu2019grid} adopts a grid localization mechanism to encode more clues for accurate object detection. 
It deals with localization as a procedure to segment grid points, which involves a heavy mask prediction process.
CenterNet~\cite{zhou2019objects} combines the classification and regression to localize the object center. It predicts possible object centers on a keypoint heatmap and then adjusts the centers by regression. A similar idea is also adopted in 3D object detection~\cite{Shi_2019_CVPR}. However, they still fall into the tradition center localization and size estimation paradigm, and the localization precision is still unsatisfactory.
LocNet~\cite{Gidaris_2016} predicts probabilities for object borders or locations inside the object's bounding box.
However, the resolution of RoI features limits the performance of LocNet because it needs to transfer the probability of pixels into the bounding box location.
On the contrary, our method focuses on the boundaries of object bounding box and decomposes the localization process for each boundary with a bucketing scheme.
We also leverage the bucketing estimation confidence to improve the classification results. Performing localization in one pass, SABL achieves substantial gains on both two-stage and single-stage pipelines while keeping their efficiency.

% !TEX root = ../2272.tex
\section{Side-Aware Boundary Localization} \label{sec:methods}
\begin{figure*}[t]
	\begin{center}
		\includegraphics[width=1.0\linewidth]{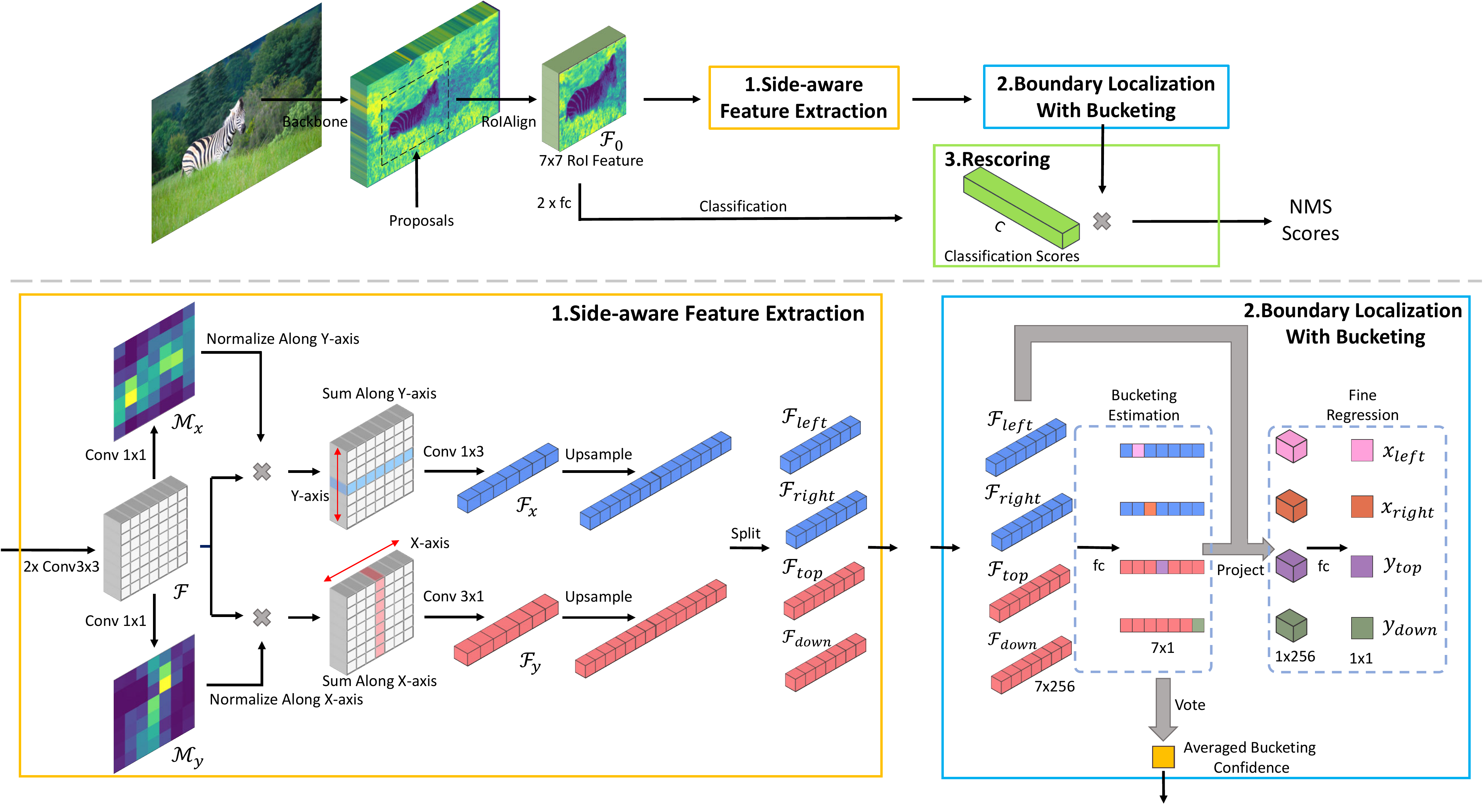}\\
	\end{center}
	\caption{
		\small{Pipeline of \textbf{Side-Aware Boundary Localization (SABL)} for the two-stage detector (see above). First, RoI features are aggregated to produce side-aware features in the Side-Aware Feature Extraction module.
			Second, the Boundary Localization with Bucketing module is performed to localize the boundaries by a two-step \emph{bucketing scheme}. Each boundary is first coarsely estimated into buckets and then finely regressed to more precise localization.
			Third, the confidences of buckets are adopted to assist the classification scores
		}
	}
	\label{fig:edge}
\end{figure*}
Accurate object localization is crucial for object detection.
Most current methods directly regress the normalized displacements between proposals and ground-truth boxes.
However, this paradigm may not provide satisfactory localization results in one pass.
Some methods~\cite{cascade_rcnn,jiang2018acquisition,wang2019region} attempt to improve localization performance with a cascading pipeline at the expense of considerable computational costs.
A lightweight as well as effective approach thus becomes necessary.

We propose Side-Aware Boundary Localization (SABL) as an alternative for the conventional bounding box regression to locate the objects more accurately.
As shown in Figure~\ref{fig:edge}, it first extracts horizontal and vertical features $\cF_x$ and $\cF_y$ by aggregating the RoI features $\cF$ along X-axis and Y-axis, respectively, and then splits $\cF_x$ and $\cF_y$ into side-aware features $\cF_{left}$, $\cF_{right}$, $\cF_{top}$ and $\cF_{down}$. (Section~\ref{sec:feature_extraction}).
Then for each side of a bounding box, SABL first divides the target space into multiple buckets (as shown in Figure~\ref{fig:motivation}) and searches for the one where the boundary resides via leveraging the side-aware features.
It will refine the boundary location $x_{left}$, $x_{right}$, $y_{top}$ and $y_{down}$ by further predicting their offsets from the bucket's centerline (Section~\ref{sec:localization}).
Such a two-step \emph{bucketing scheme} could reduce the regression variance and ease the difficulties of prediction.
Furthermore, the confidence of estimated buckets could also help to adjust the classification scores and further improve the performance (Section~\ref{sec:rescoring}).
With minor modifications, SABL is also applicable for single-stage detectors (Section~\ref{sec:single_stage}).
\subsection{Side-Aware Feature Extraction}\label{sec:feature_extraction}
As shown in Figure~\ref{fig:edge}, we extract side-aware features $\cF_{left}$, $\cF_{right}$, $\cF_{top}$, and $\cF_{down}$ based on the $k\times k$ RoI features $\cF$ ($k = 7$).
Following typical conventions~\cite{girshick2015fast,ren2015faster,mask_rcnn}, we adopt RoIAlign to obtain the RoI feature of each proposal. 
Then we utilize two $3\times3$ convolution layers to transform it to $\cF$.
To better capture direction-specific information of the RoI region, we employ the self-attention mechanism to enhance the RoI feature. Specifically, we predict two different attention maps from $\cF$ with a $1\times 1$ convolution, which are then normalized along the Y-axis and X-axis, respectively.
Taking the attention maps $\cM_x$ and $\cM_y$, we aggregate $\cF$ to obtain $\cF_x$ and $\cF_y$ as follows,
{\small{
\begin{equation} \label{eq:kernel-prediction}
	\begin{split}
	\cF_x = \sum_{y}\cF(y, :)*\cM_x(y, :), \\
	\cF_y = \sum_{x}\cF(:, x)*\cM_y(:, x).
	\end{split}
\end{equation}
}
}
$\cF_x$ and $\cF_y$ are both a 1-D feature map of shape $1\times k$ and $k\times 1$, respectively.
They are further refined by a $1\times3$ or $3\times1$ convolution layer and upsampled by a factor of 2 through a deconvolution layer, resulting in $1\times 2k$ and $2k\times 1$ features on the horizontal and vertical directions, respectively.
Finally, the upsampled features are simply split into two halves, leading to the side-aware features $\cF_{left}$, $\cF_{right}$, $\cF_{top}$ and $\cF_{down}$.

\subsection{Boundary Localization with Bucketing}\label{sec:localization}

As shown in the module 2 of Figure~\ref{fig:edge}, we decompose the localization process into a two-step \emph{bucketing scheme}: bucketing estimation and fine regression.
The candidate region of each object boundary is divided into buckets horizontally and vertically.
We first estimate in which bucket the boundary resides and then regress a more accurate boundary localization from this bucket.

\noindent\textbf{Two-Step Bucketing Scheme.}
Given a proposal box, \ie, $(B_{left},B_{right},B_{top},$
$B_{down})$, we relax the candidate region of boundaries by a scale factor of $\sigma$ ($\sigma>1$), to cover the entire object.
The candidate regions are divided into $2k$ buckets on both X-axis and Y-axis, with $k$ buckets corresponding to each boundary.
The width of each bucket on X-axis and Y-axis are therefore $l_x=(\sigma B_{right} - \sigma B_{left}) / 2k$ and $l_y = (\sigma B_{down} - \sigma B_{top}) / 2k$, respectively.
\begin{SCfigure}[][t]
	\centering
	\includegraphics[width=0.65\linewidth]{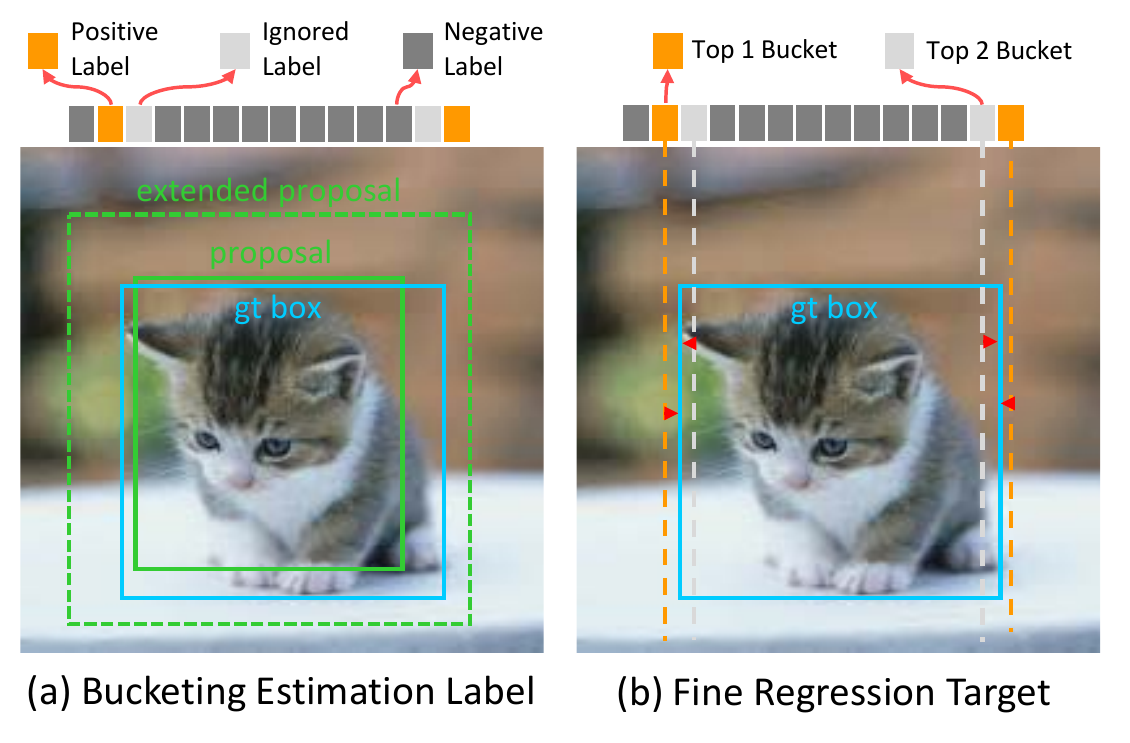}
	\caption{\small{The localization target of \textbf{SABL} for bucketing estimation and fine regression on X-axis. The localization target for Y-axis can be calculated similarly}
	}
	\label{fig:target}
\end{SCfigure}
In the bucketing estimation step, we adopt a binary classifier to predict whether the boundary is located in or is the closest to the bucket on each side, based on the side-aware features.
In the fine regression step, we apply a regresser to predict the offset from the centerline of the selected bucket to the ground-truth boundary.

\noindent
\textbf{Localization Targets.}
There are a bucketing estimation and a fine regression branch in the \emph{bucketing scheme} to be trained.
We follow the conventional methods~\cite{girshick2015fast,ren2015faster} for label assigning and proposal sampling.
The bucketing estimation determines the nearest buckets to the boundaries of a ground-truth bounding box by binary classification. 
As shown in Figure~\ref{fig:target}, on each side, the bucket, whose centerline is the nearest to the ground-truth boundary, is labeled as 1 (positive sample), while the others are labeled as 0 (negative samples).
To reduce the ambiguity in training, on each side, we ignore the bucket that is the second nearest to the ground-truth boundary because it is hard to be distinguished from the positive one.
For each side, we ignore negative buckets when training the boundary regressor.
To increase the robustness of the fine regression branch, we include both the nearest (labeled as ``positive'' in the bucketing estimation step) bucket and the second nearest (labeled as ``ignore'' in the bucketing estimation step) bucket to train the regressor.
The regression target is the displacement between the bucket centerline and the corresponding ground-truth boundary.
To ease the training difficulties of regressors, we normalize the target by $l_x$ and $l_y$ on the corresponding axes.

\subsection{Bucketing-Guided Rescoring}\label{sec:rescoring}
The bucketing scheme brings a natural benefit, \ie, the bucketing estimation confidences can represent the reliability of predicted locations.
With the aim at keeping the more accurately localized bounding boxes
during non-maximal suppression (NMS),
we utilize the localization reliability to guide the rescoring.
Therefore, SABL averages the bucketing estimation confidence scores of four boundaries. The multi-category classification scores are multiplied by the averaged localization confidence, and then used for ranking candidates during NMS.
The rescoring helps maintain the best box with both high classification confidence and accurate localization.

\begin{figure}[t]
	\begin{center}
		\includegraphics[width=0.86\linewidth]{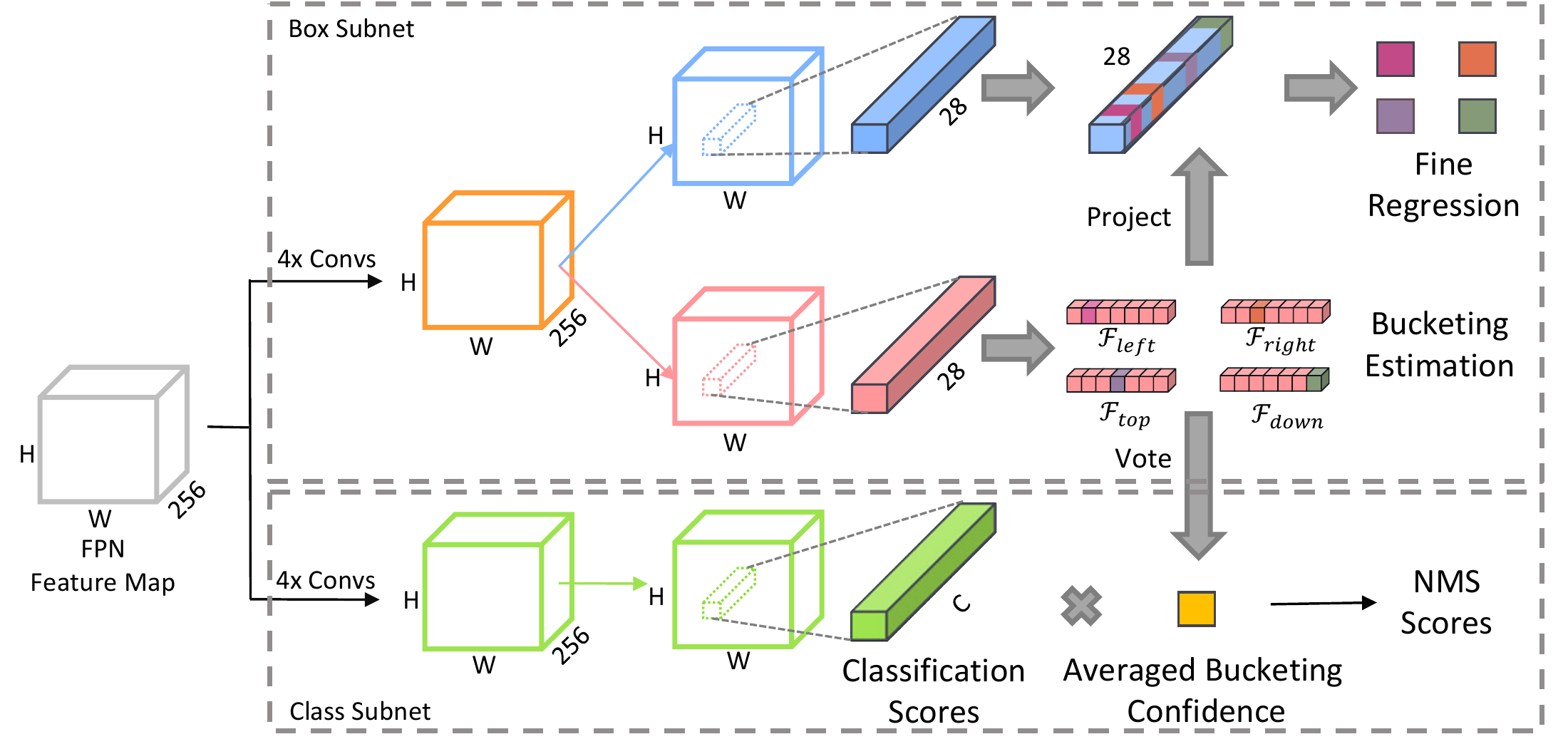}\\
	\end{center}
	\caption{
		\small{Pipeline of \textbf{Side-Aware Boundary Localization (SABL)} for the single-stage detector. Since there is no RoI features, SABL adopts convolution layers to produce the feature for localization at each location.
			Then bucketing estimation and fine regression are performed based on this feature at each location.
			Furthermore, bucketing estimation confidence is leveraged to adjust the classification scores as well}
	}
	\label{fig:edge_single}
\end{figure}

\subsection{Application to Single-Stage Detectors}\label{sec:single_stage}
SABL can also be applied to single-stage detectors such as~\cite{lin2017_focal}, with minor modifications.
Since there is no proposal stage in single-stage detectors, Side-Aware Feature Extraction (SAFE) is not adopted and the feature extraction is performed following RetinaNet~\cite{lin2017_focal}.
As shown in Figure~\ref{fig:edge_single}, on top of the FPN features, four convolution layers are adopted to classification and localization branches respectively.
Following the state of the arts~\cite{tian2019fcos,zhou2019objects,kong2019foveabox}, Group Normalization (GN)~\cite{wu2018group} is adopted in these convolution layers.
At each position of FPN feature maps, there is only one anchor used for detection following~\cite{wang2019region}.
The size of this anchor is $\gamma * s$, where $\gamma$ is a hyperparameter ($\gamma=8$), and $s$ is the stride of the current feature map.
SABL learns to predict and classify one bounding box based on this anchor.
The target assignment process follows the same setting as in~\cite{wang2019region}. Specifically, we utilize multiple (9 by default) anchors on each location to compute IoUs and match the ground-truths for this location during training, but the forward process only involves one anchor per location.
This design enables SABL to cover more ground-truths and be better optimized, as well as keeping its efficiency.
After using convolution layers to produce the feature for localization on each position, the ensuing Boundary Localization with Bucketing and Bucketing-Guided Rescoring remain the same.

% !TEX root = ../2272.tex

\section{Experiments}
\subsection{Experimental Setting}
\noindent
\textbf{Dataset.}
We perform experiments on the challenging MS COCO 2017 benchmark~\cite{lin2014coco}.
We use the \emph{train} split for training and report the performance
on the \emph{val} split for ablation study.
Detection results for comparison with other methods are reported on the \emph{test-dev} split if not further specified.

\noindent
\textbf{Implementation Details.}
During training, We follow the 1x training scheduler~\cite{Detectron2018} and use mmdetection \cite{mmdetection} as the codebase. We train Faster R-CNN~\cite{ren2015faster}, Cascade R-CNN~\cite{cascade_rcnn} and RetinaNet~\cite{lin2017_focal} with batch size of 16 for 12 epochs.
We apply an initial learning rate of 0.02 for Faster R-CNN, Cascade R-CNN, and 0.01 for RetinaNet. ResNet-50~\cite{He_2016} with FPN~\cite{lin2017_fpn} backbone is adopted if not further specified.
The long edge and short edge of images are resized to 1333 and 800 respectively without changing the aspect ratio during training and inference if not otherwise specified.
The scale factor $\sigma$ is set as 1.7 and 3.0 for Faster R-CNN and RetinaNet, respectively. For Cascade R-CNN, we replace the original bbox head with the proposed SABL, and $\sigma$ for three cascading stages are set as 1.7, 1.5 and 1.3, respectively. $k$ is set to 7 for all experiments if not further specified.
GN is adopted in RetinaNet and RetinaNet w/ SABL as in Sec~\ref{sec:single_stage} but not in Faster R-CNN and Cascade R-CNN. Detection results are evaluated with the standard COCO metric.
The runtime is measured on a single Tesla V100 GPU.

\noindent
\textbf{Training Details.}
The proposed framework is optimized in an end-to-end manner.
For the two-stage pipeline, the RPN loss $\cL_{rpn}$ and classification loss $\cL_{cls}$ remain the same as Faster R-CNN~\cite{ren2015faster}.
We replace the bounding box regression loss by a bucketing estimation loss $\cL_{bucketing}$ and a fine regression loss $\cL_{reg}$.
Specifically, $\cL_{bucketing}$ adopts Binary Cross-Entropy Loss, $\cL_{reg}$ applies Smooth L1 Loss.
In summary, a general loss function can be written as follows: 
$\cL = \lambda_1\cL_{rpn} + \cL_{cls} + \lambda_2(\cL_{bucketing} + \cL_{reg})$,
where $\lambda_1 = 1, \lambda_2 = 1$ for the two-stage pipeline,  $\lambda_1 = 0, \lambda_2 = 1.5$ for the single stage
pipeline.

\subsection{Results}
We show the effectiveness of SABL by applying SABL on RetinaNet, Faster R-CNN and Cascade R-CNN with ResNet-101~\cite{He_2016} with FPN~\cite{lin2017_fpn} backbone.
To be specific, we adopt SABL to Faster R-CNN and RetinaNet as described in Sec.~\ref{sec:methods}.
As shown in Table~\ref{tab:results}, SABL improves the $AP$ of RetinaNet by $1.7\%$ with no extra cost, and Faster R-CNN by $3.0\%$ with only around 10\% extra inference time. We further apply SABL to the powerful Cascade R-CNN. SABL improves the performance by $0.9\%$ on this strong baseline. 

\begin{table*}[ht]
	\caption{
		\small{Comparison to mainstream methods with ResNet-101 FPN backbone on COCO dataset. \emph{m.s.} indicates multi-scale training. \emph{Sch.} indicates training schedule. 50e indicates 50 epochs. \emph{Data} indicates the results are evaluated on the corresponding data split of COCO dataset, \eg,  some two-stage detectors are evaluated on COCO val split}
	}
	\label{tab:results}
	\begin{center}
		\begin{tabular}{c|c|c|ccccccc}
			\hline
			Method & Backbone & Sch. & AP & $AP_{50}$ & $AP_{75}$ & $AP_{S}$ & $AP_{M}$ & $AP_{L}$ & FPS \\
			%&  &  &  &  &  &  &  & & \\ 
			\hline
			RetinaNet~\cite{lin2017_focal} & ResNet-101 & 1x & 38.8 & 60.0 & 41.7 & 21.9 & 42.1 & 48.6 & 13.0 \\
			FSAF~\cite{Zhu_2019_CVPR} (m.s.) & ResNet-101 & 1.5x & 40.9 & 61.5 & 44.0 & 24.0 & 44.2 & 51.3 & 12.4\\
			FCOS~\cite{tian2019fcos} (m.s.) & ResNet-101 & 2x & 41.5 & 60.7 & 45.0 & 24.4 & 44.8 & 51.6 & 13.5 \\
			GA-RetinaNet~\cite{wang2019region} (m.s.) & ResNet-101 & 2x  & 41.9 & 62.2 & 45.3 & 24.0 & 45.3 & 53.8 & 11.7\\
			CenterNet~\cite{zhou2019objects} (m.s.) & Hourglass-104 & 50e & 42.1 & 61.1 & 45.9 & 24.1 & 45.5 & 52.8 & 8.9\\
			FoveaBox~\cite{kong2019foveabox} (m.s.) & ResNet-101 & 2x  & 42.0 & 63.1 & 45.2 & 24.7 & 45.8 & 51.9 & 12.8\\
			RepPoints~\cite{yang2019reppoints} (m.s.) & ResNet-101 & 2x  & 42.6 & 63.5 & 46.2 & 25.4 & 46.2& 53.3 & 12.2 \\
			\hline
			RetinaNet w/ SABL & ResNet-101 & 1x & 40.5 & 59.3 & 43.6 & 23.0 & 44.1 & 51.3 &  13.0 \\
			RetinaNet w/ SABL (m.s.) & ResNet-101 & 1.5x & 42.7 & 61.4 & 46.0 & 25.3 & 46.8 & 53.5 & 13.0 \\ 
			RetinaNet w/ SABL (m.s.) & ResNet-101 & 2x & \textbf{43.2} & 62.0 & 46.6 & 25.7 & 47.4 & 53.9 &  13.0 \\ \hline
		\end{tabular}
	\end{center}
	\addtolength\tabcolsep{-0.1em}
	\begin{center}
		\begin{tabular}{c|c|c|ccccccc}
			\hline
			Method & Backbone & Data & AP & $AP_{50}$ & $AP_{75}$ & $AP_{S}$ & $AP_{M}$ & $AP_{L}$ & FPS \\
			%&  &  &  &  &  &  &  & & \\ 
			\hline
			Faster R-CNN~\cite{ren2015faster} & ResNet-101 & val &38.5  &60.3 & 41.6 & 22.3 & 43.0 & 49.8 & 13.8 \\
			Faster R-CNN~\cite{ren2015faster} & ResNet-101 & test-dev &38.8  &60.9 & 42.3 & 22.3 & 42.2 & 48.6 & 13.8 \\
			IoU-Net~\cite{jiang2018acquisition} & ResNet-101 & val & 40.6 & 59.0 & - & - & - & - & - \\
			GA-Faster R-CNN~\cite{wang2019region} & ResNet-101 & test-dev & 41.1 & 59.9 & 45.2 & 22.4 & 44.4 & 53.0 & 11.5 \\
			Grid R-CNN Plus~\cite{lu2019grid_plus} & ResNet-101& test-dev & 41.4 &60.1 & 44.9 & 23.4 & 44.8 & 52.3 & 11.1 \\
			\hline
			Faster R-CNN w/ SABL &ResNet-101 & val &41.6 & 59.5 & 45.0 & 23.5 & 46.5 & 54.6 & 12.4 \\
			Faster R-CNN w/ SABL &ResNet-101 & test-dev & \textbf{41.8} &60.2 & 45.0 & 23.7 & 45.3 & 52.7 & 12.4 \\
			\hline
			\hline
			Cascade R-CNN~\cite{cascade_rcnn} & ResNet-101& test-dev & 42.4 &61.1 & 46.1 & 23.6 & 45.4 & 54.1 & 11.2 \\
			\hline
			\textbf{Cascade R-CNN w/ SABL} & ResNet-101 & test-dev & \textbf{43.3} & 60.9  & 46.2 & 23.8 & 46.5 & 55.7 & 8.8 \\ 
			\hline
		\end{tabular}
	\end{center}
\end{table*}

The significant performance gains on various object detection architectures show that SABL is a generally efficient and effective bounding box localization method for object detection.
We further compare SABL with other advanced detectors in Table~\ref{tab:results}. The reported performances here either come from the original papers or from released implementations and models.
SABL exhibits the best performance among these methods and retains its efficiency. To make a fair comparison with other single stage-detectors, we employ multi-scale training, \ie, randomly scaling the shorter edge of input images from 640 to 800 pixels,
and the training schedule is extended to 2x. For two-stage detectors, the 1x training schedule is adopted. 
As shown in Table~\ref{tab:results}, Faster R-CNN w/ SABL outperforms recent two-stage detectors~\cite{jiang2018acquisition,wang2019region,lu2019grid_plus} that also aim at better localization precision. To be specific, IoU-Net~\cite{jiang2018acquisition} and GA-Faster RCNN~\cite{wang2019region} adopt iterative regression, and Grid R-CNN Plus~\cite{lu2019grid_plus} improves localization by leveraging a keypoint prediction branch. 
The experimental results reveal the advantages of the proposed SABL among advanced localization pipelines.

\begin{table}[t]
	\centering
	\caption{\small{The effects of each module in our design. \emph{SAFE}, \emph{BLB}, \emph{BGR} denote Side-Aware Feature Extraction, Boundary Localization with Bucketing and Bucketing-Guided Rescoring, respectively}}
	\begin{center}
		\begin{tabular}{ccc|ccccccc}
			\hline
			SAFE & BLB & BGR & $AP$ & $AP_{50}$ & $AP_{75}$ & $AP_{90}$ & $AP_{S}$ & $AP_{M}$ & $AP_{L}$ \\ \hline
			 &   &    &  36.4 & 58.4 & 39.3 & 8.3 & 21.6 & 40.0 & 47.1 \\
			\checkmark &  &    & 38.5 & 58.2 & 41.6  & 14.3 & 23.0 & 42.5 & 49.5 \\
			 & \checkmark &    & 38.3 & 57.6 & 40.5 & 16.1 & 22.3 & 42.6 & 49.7 \\
			 & \checkmark & \checkmark &  39.0 & 57.5 & 41.9  & 17.1 & 22.6 & 43.2 & 50.9 \\
			\checkmark & \checkmark &  &  39.0 & 57.9 & 41.4  & 17.8 & 22.7 & 43.4 & 49.9 \\
			\checkmark & \checkmark & \checkmark  & 39.7 & 57.8 & 42.8 & 18.8 & 23.1 & 44.1 & 51.2 \\ \hline
		\end{tabular}
	\end{center}
	\label{tab:components}
\end{table}

\subsection{Ablation Study}

\noindent
\textbf{Model Design.}
We omit different components of SABL on two-stage pipeline to investigate the effectiveness
of each component, including Side-Aware Feature Extraction (SAFE), Boundary Localization with Bucketing (BLB) and Bucketing-Guided Rescoring (BGR). The results are shown in Table~\ref{tab:components}.
We use Faster R-CNN with ResNet-50~\cite{He_2016} w/ FPN~\cite{lin2017_fpn} backbone as the baseline. Faster R-CNN adopts center offsets and scale factors of spatial sizes, \ie, $(\delta x, \delta y, \delta w, \delta h)$ as regression targets and achieves 36.4\% $AP$ on COCO val set. SABL significantly improves the baseline by $3.3\%$ AP, especially on high IoU thresholds, \eg, SABL tremendously improves $AP_{90}$ by $10.5\%$.

\emph{Side-Aware Feature Extraction (SAFE)}. In Table~\ref{tab:components}, we apply Side-Aware Feature Extraction (SAFE) as described in Sec.~\ref{sec:feature_extraction}.
In order to leverage the side-aware features, side-aware regression targets are required. We introduce \emph{boundary regression} targets, \ie, the offset of each boundary $(\delta x_1, \delta y_1, \delta x_2, \delta y_2)$. This simple modification improves the performance from 36.4\% to 37.3\%, demonstrating that localization by each boundary is more preferable than regressing the box as a whole.
SAFE focuses on content of the corresponding side and further improves the performance from 37.3\% to 38.5\%.
To verify that simply adding more parameters will not apparently improve the performance,
we also train a Faster RCNN with \emph{boundary regression} and 4conv1fc head. The 4conv1fc head contains four 3x3 convolution layers followed by one fully-connected layer. Although the 4conv1fc head is heavier than SAFE, it marginally improves the $AP$ by 0.1\%, \ie, from 37.3\% to 37.4\%.

\emph{Boundary Localization with Bucketing (BLB)}. 
As described in Sec.~\ref{sec:localization}, BLB divides the RoI into multiple buckets, it first determines which bucket the boundary resides and takes the centerline of the selected bucket as a coarse estimation of boundary.
Then it performs fine regression to localize the boundary precisely. BLB achieves 38.3\%, outperforming the popular bounding box regression by 1.9\%. Combining BLB with SAFE further improves the $AP$ to 39.0\%.

\emph{Bucketing-Guided Rescoring (BGR)}. 
Bucketing-Guided Rescoring (BGR) is proposed to adjust the classification scores as in Sec.~\ref{sec:rescoring}. 
The bucketing confidence can naturally be used to represent how confident the model believes that a boundary is precisely localized.
We average the confidences of selected buckets for four boundaries and multiply it to classification scores before NMS.
Applying the BGR further improves the performance by 0.7\% $AP$.

\noindent\textbf{Side-Aware Feature Extraction.}
Side-Aware Feature Extraction (SAFE) is used to aggregate the 2D RoI features to 1D features for X-axis and Y-axis, respectively. Here we perform a thorough ablation study for SAFE.

\begin{table}[t]
	\caption{
		\small{Number of convolution layers for Side-Aware Feature Extraction (SAFE) module.
			\emph{2D Conv} indicates the number of 3x3 Convolution layers before $\cF$.
			\emph{1D Conv} indicates the number of 1x3 and 3x1 convolution layers before $\cF_x$ and $\cF_y$, respectively
		}
	}
	\label{tab:feature_conv_nums}
	\begin{center}
		\begin{tabular}{ccccc||ccccc}
			\hline
			2D Conv & 1D Conv & AP & Param & FLOPS & 2D Conv & 1D Conv & AP  & Param & FLOPS \\ \hline
			0 & 1 & 38.3 & 40.8M & 212G & 2 & 0 & 39.5 & 41.6M & 267G \\
			0 & 2 & 38.3 & 41.2M & 215G & \textbf{2} & \textbf{1} & 39.7 & 42M & 270G \\
			1 & 1 & 39.3 & 41.4M & 241G & 2 & 2 & 39.6 & 42.4M & 273G \\
			1 & 2 & 39.4 & 41.8M & 244G & 3 & 1 & 39.7 & 42.6M & 299G \\
			\hline
		\end{tabular}
	\end{center}
\end{table}

\emph{Parameters.} In SAFE, after performing the RoI pooling, we apply two 3x3 convolution layers to obtain $\cF$. We adopt one 1x3 and the other 3x1 convolution layers after aggregating the 1D features on horizontal and vertical directions to obtain $\cF_x$ and $\cF_y$, respectively.
We investigate the influence of these convolution layers. As shown in Table~\ref{tab:feature_conv_nums}, we list the performance as well as parameters and FLOPS under different settings. 
It's noteworthy that Faster R-CNN w/ SABL still achieves satisfactory performance with smaller computational cost.
Thus the proposed method could be flexibly adjusted to fulfill different requirements of computational cost.

\begin{table}[tb]
	\begin{minipage}[t]{0.475\linewidth}
			\caption{
				\small{Comparison of different methods to aggregate the 2D RoI features into 1D features in SAFE module}
			}
			\label{tab:feature_attention}
			\begin{center}
				\begin{tabular}{c|c}
					\hline
					Aggregating Method & $AP$ \\ \hline
					Max Pooling & 39.4  \\
					Average Pooling & 39.3 \\ 
					Attention Mask & 39.7 \\ \hline
				\end{tabular}
			\end{center}
	\end{minipage}
\hfill
	\begin{minipage}[t]{0.475\linewidth}
			\caption{
				\small{Comparison of different settings of feature size in SAFE module}
			}
			\label{tab:feature_size}
			\begin{center}
				\begin{tabular}{cc|cc}
					\hline
					RoI & Upsample & $AP$ & FLOPS \\ \hline
					7 & 7 & 39.0 & 266G\\
					\textbf{7} & \textbf{14} & 39.7 & 270G \\
					7 & 28 & 39.1 & 281G \\
					14 & 14 & 39.7 & 443G \\ \hline
				\end{tabular}
			\end{center}
	\end{minipage}
\end{table}

\emph{Feature aggregating method}. 
As in Sec~\ref{sec:feature_extraction}, we apply a self-attention mechanism to aggregate 2D RoI features into 1D features.
The max pooling and average pooling are two alternative
approaches in this procedure. In Table~\ref{tab:feature_attention}, experimental results reveal that the proposed attention mask is more effective than max or average pooling to aggregate RoI features.

\emph{Size of Side-Aware Features}. In the Side-Aware Feature Extraction module, we first perform RoI-Pooling and get the RoI features with spatial size $7 \times 7$. The RoI features are aggregated into 1D features with size $1 \times 7$ and $7 \times 1$, and then upsampled to size of $1 \times 14$ and $14 \times 1$ by a deconvolution layer. We study RoI features size and upsampled features size. As shown in Table~\ref{tab:feature_size}, our settings, \ie, RoI size of 7 and upsampled size of 14, achieve the best trade-off between effectiveness and efficiency.

\noindent\textbf{Boundary Localization with Bucketing.} Here we discuss the effectiveness of different designs for localization. In our work, we propose Boundary Localization with Bucketing (BLB), that contains 3 key ideas, \ie, localizing by boundaries, bucketing estimation and fine regression.
\begin{table}[t]
	\caption{
		\small{Influence of different localization pipelines. To crystallize the effectiveness of Boundary Localization of Bucketing (BLB), Side-Aware Feature Extraction (SAFE) and Bucketing Guided Rescoring (BGR) are not applied here}
	}
	\label{tab:BLB}
	\small
	\begin{center}
		\begin{tabular}{c|ccccccc}
			\hline
			Localization Approach &$AP$ & $AP_{50}$ & $AP_{75}$ & $AP_{90}$ & $AP_{S}$ & $AP_{M}$ & $AP_{L}$ \\ \hline
			Bounding Box Regression & 36.4 & 58.4 & 39.3 & 8.3 & 21.6 & 40.0 & 47.1 \\
			Boundary Regression & 37.3 & 58.2 & 40.4 & 10.6 & 22.0 & 41.2 & 47.8 \\
			Bucketing & 32.8 & 56.7 & 35.9 & 2.0 & 20.1 & 36.5 & 41.5 \\
			Iterative Bucketing & 36.8 & 58.3 & 40.9 & 6.0 & 20.8 & 40.2 & 48.1 \\
			Center Localization with Bucketing (CLB) & 36.9 & 57.6 & 39.5 & 11.4 & 20.8 & 41.2 & 47.7 \\
			Boundary Localization with Bucketing (BLB) & 38.3 & 57.6 & 40.5 & 16.1 & 22.3 & 42.6 & 49.7 \\ \hline
		\end{tabular}
	\end{center}
\end{table}
\begin{SCfigure}[][t]
	\centering
	\includegraphics[width=0.7\linewidth]{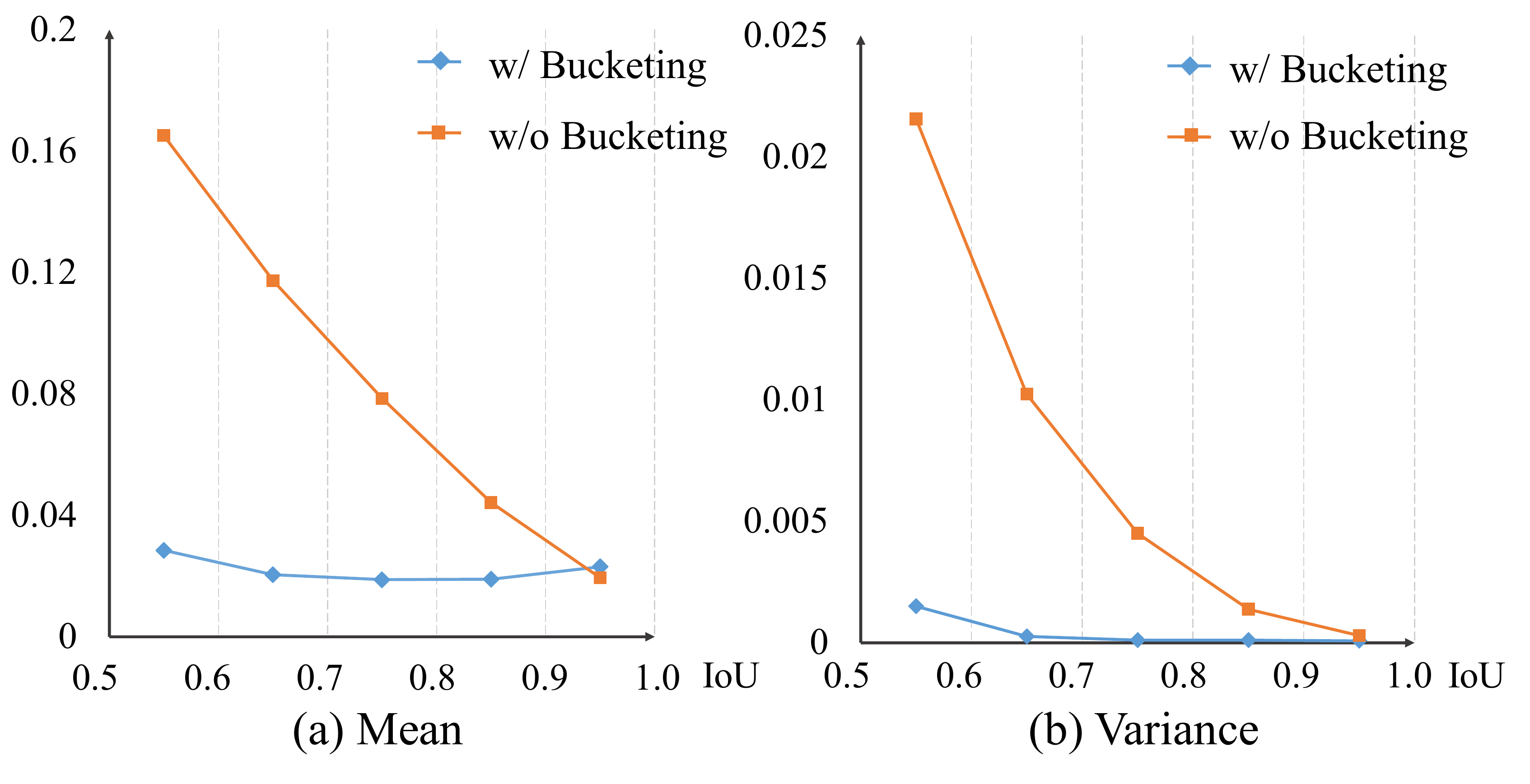}
	\caption{
		\small{Mean and variance of displacements from proposals to ground-truth boundaries \wrt the size of the ground-truth boxes with or without bucketing}
	}
	\label{fig:displacement}
\end{SCfigure}

As shown in Table~\ref{tab:BLB}, the proposed BLB achieves significantly higher performance than the widespread \emph{Bounding Box Regression} (38.3\% vs.\ 36.4\%), that adopts center offsets and scale factors of spatial sizes, \ie, $(\delta x, \delta y, \delta w, \delta h)$ as regression targets.
Switching to \emph{Boundary Regression} that regresses boundary offsets $(\delta x_1, \delta y_1, \delta x_2, \delta y_2)$, improves the $AP$ by $0.9\%$.
The result reveals that localizing the object boundaries is more preferable than localizing object centers.
Moreover, to show the advantages of the proposed design to iterative \emph{Bounding Box Regression}, we compare SABL with IoUNet~\cite{jiang2018acquisition}, GA-Faster R-CNN~\cite{wang2019region}, GA-RetinaNet~\cite{wang2019region} in Table~\ref{tab:results}.

\emph{Bucketing} indicates adopting the centerline of the predicted bucket for each boundary as the final localization. It presents a much inferior performance.
Due to the absence of fine regression, the localization quality is severely affected by the bucket width.
Following LocNet~\cite{Gidaris_2016}, we design a heavy \emph{Iterative Bucketing} where the bucketing step is performed iteratively. 
Although the performance is improved from $32.8\%$ to $36.8\%$, it remains inferior to 38.3\% of our method.

We also investigate a scheme to localize the object center with bucketing named \emph{Center Localization with Bucketing (CLB)}.
Bucketing estimation and fine regression are used to localize the object center, and width and height are then regressed as in the conventional \emph{Bounding Box Regression}.
CLB achieves $1.4\%$ lower $AP$ than BLB, which further validates the necessity of localizing object boundaries other than the center.

\begin{table}[h]
	\begin{minipage}[t]{0.38\linewidth}
		\caption{
			\small{Influence of different designs to generate regression targets. \emph{Ignore} and \emph{Top2-Reg} are described in \emph{Target design}}
		}
		\label{tab:taget_design}
		\begin{center}
			\begin{tabular}{cc|c}
				\hline
				Ignore & Top2-Reg & $AP$ \\ \hline
				&  & 38.7  \\
				\checkmark&  & 39.1  \\
				&  \checkmark& 39.4 \\
				\checkmark&  \checkmark& 39.7  \\ \hline
			\end{tabular}
		\end{center}
	\end{minipage}
	\hfill
	\begin{minipage}[t]{0.6\linewidth}
		\caption{
			\small{Influence of different hyper-parameters in RetinaNet w/ SABL, \ie, scale factor $\sigma$, buckets number and localization loss weight $\lambda_2$. GN is not adopted in this table}
		}
		\label{tab:singlestage}
		\begin{center}
			\begin{tabular}{cccc||cccc}
				\hline
				$\sigma$ & Bucket-Num & $\lambda_2$ & AP & $\sigma$ & Bucket-Num & $\lambda_2$ & AP \\ \hline
				2 & 7 & 1.5 & 37.3 & 3 & 9 & 1.5 & 37.2 \\
				\textbf{3} & \textbf{7} & \textbf{1.5} & 37.4 & 3 & 7 & 1.0 & 36.8 \\
				4 & 7 & 1.5 & 36.9 & 3 & 7 & 1.25 & 37.2 \\
				3 & 5 & 1.5 & 36.9 & 3 & 7 & 1.75 & 37.2 \\
				\hline
			\end{tabular}
		\end{center}
	\end{minipage}
\end{table}

Figure~\ref{fig:displacement} shows mean and variance of displacements from proposals to ground-truth boxes which are normalized by the size of ground-truth boxes.
Without loss of generality, we choose the left boundary to calculate the statistic. The proposals are split into five groups according to their IoU with the ground-truth, \ie, [0.5, 0.6), [0.6, 0.7), [0.7, 0.8), [0.8, 0.9), [0.9, 1.0). 
Regression with bucketing exhibits more stable distribution on displacements, easing the difficulties of regression and lead to more precise localization. 

\emph{Target design}. We further study the training target designs for this module.
1) \emph{Ignore}: During training bucketing estimation branch, we ignore the second nearest bucket to ease its ambiguity with the nearest bucket. 
2) \emph{Top2-Reg}: During training the fine regression branch, buckets with Top-2 displacements to the ground-truth boundaries are trained to reduce the influence of mis-classification in bucketing estimation.
As shown in Table~\ref{tab:taget_design}, two proposed designs bring substantial performance gains. In our study, the classification accuracy of Top-1, Top-2 and Top-3 buckets are 69.3\%, 90.0\% and 95.7\%, respectively. 
We also try to train with Top-3 regression targets, however the performance remains 39.7\%.

\emph{Scale factor}.
We study the influence of different scale factors $\sigma$ to enlarge proposals during generating localization targets. 
To be specific, when adopting $\sigma$ of 1.1, 1.3, 1.5, \textbf{1.7}, 1.9, the performance are 39.2\%, 39.4\%, 39.6\%, \textbf{39.7\%}, 39.6\%, respectively.

\noindent\textbf{SABL for Single-Stage Detectors.}
For single-stage detectors, we take RetinaNet~\cite{lin2017_focal} as a baseline. Following conventions of recent single-stage methods~\cite{tian2019fcos,zhou2019objects,kong2019foveabox}, GN is adopted in the head of both RetinaNet and RetinaNet w/ SABL.
GN improves RetinaNet from 35.6\% to 36.6\% and RetinaNet w/SABL from 37.4\% to 38.5\%. SABL shows consistent improvements over the baseline.
Since the single-stage pipeline is different from the two-stage one, we study the hyper-parameters as shown in Table~\ref{tab:singlestage}.
Results reveal that the setting of $\sigma=3,\lambda_2=1.5$ and a bucket number of 7 achieves the best performance.

\begin{figure}[t]
	\centering
	\includegraphics[width=0.83\linewidth]{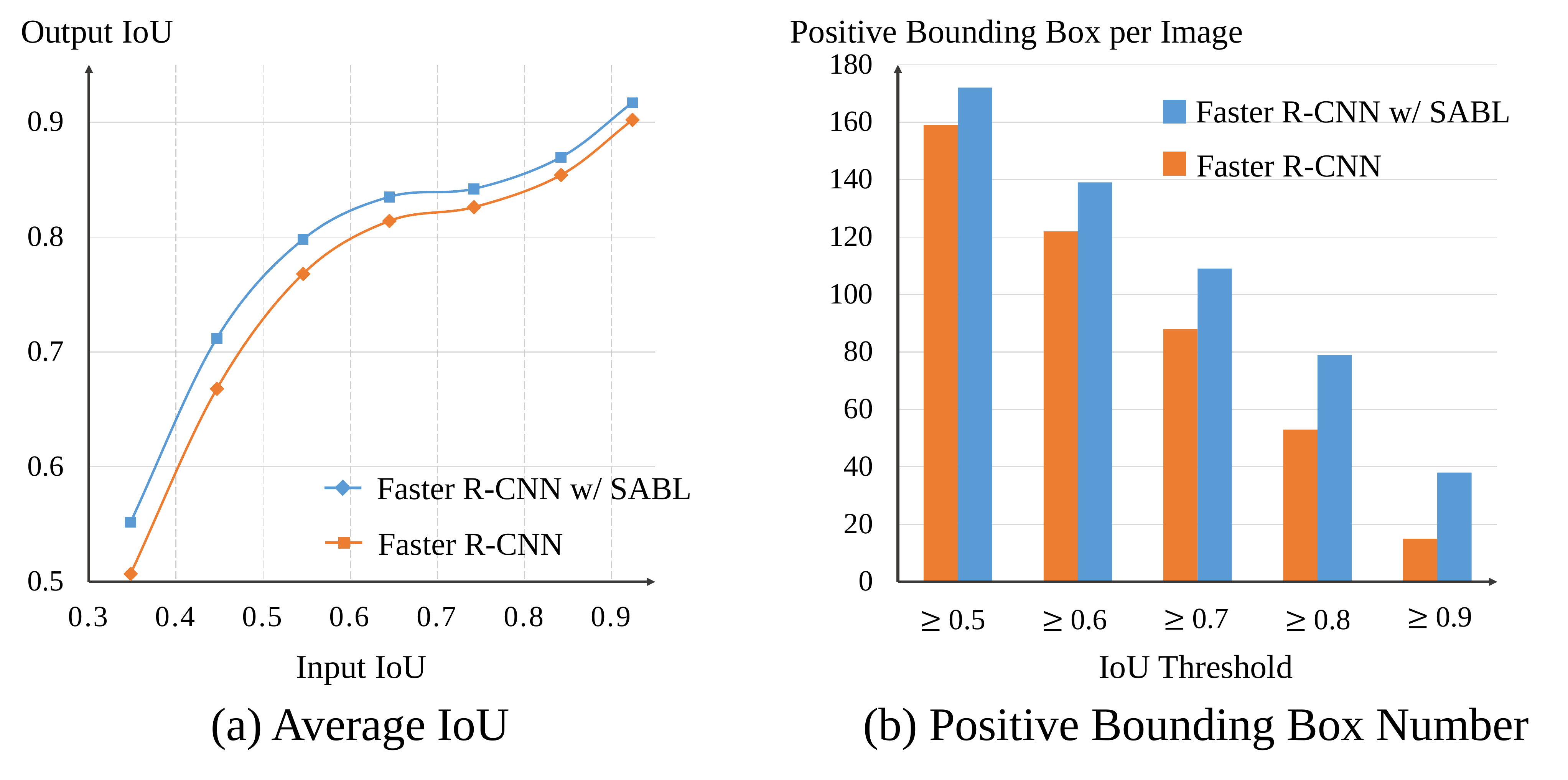}
	\caption{
		\small{Analysis of bounding boxes predicted by Faster R-CNN and Faster R-CNN w/ SABL without NMS. (a) Average IoU of proposals before and after localization branch. (b) Number of positive boxes per image with different IoU threshold after localization }
	}
	\label{fig:analysis}
\end{figure}

\noindent\textbf{Analysis of Localization Precision.}
To demonstrate the effectiveness of SABL on improving the localization quality, we perform quantitative analysis on Faster R-CNN and Faster R-CNN w/ SABL.
We split the proposals into different bins ($[0.3, 0.4), [0.4, 0.5),\dots,[0.9, 1)$) according to the IoUs with their nearest ground-truth object, and then compare the average IoU before and after the localization branch in each bin.
As shown in Figure~\ref{fig:analysis} (a), SABL achieves consistently higher IoU than the bounding box regression baseline in all bins, which reveals that both low and high quality proposals are more precisely localized.

Furthermore, in Figure~\ref{fig:analysis} (b) we compare the IoU distribution of proposals after the localization branch.
Specifically, we calculate the average number of positive boxes per image with different IoU threshold (\eg, IoU $\geq$ 0.5).
SABL results in more positive boxes under all thresholds, especially for high IoU thresholds, \eg, $\geq$ 0.9.
It contributes to the significant gains of Faster R-CNN w/ SABL on $AP_{90}$ compared to Faster R-CNN.
We also notice that although SABL achieves a higher overall AP and better localization precision across all IoU thresholds, $AP_{50}$ is slightly lower.
The situation of higher overall $AP$ but lower $AP_{50}$, also occurs in a number of detectors~\cite{jiang2018acquisition,lu2019grid,wang2019region} that aim at better localization.
AP is affected by not only the localization quality but classification accuracy, and $AP_{50}$ is more sensitive to classification since it does not require bounding boxes with high IoU.
Localization and classification branches are jointly trained, and SABL is more optimized for the former.
To improve $AP_{50}$, other efforts to obtain a higher classification accuracy are required, \eg, reducing misclassified boxes, which is beyond the discussion and target of our method.

% !TEX root = ../2272.tex
\section{Conclusion}
In this work, we propose \textbf{Side-Aware Boundary Localization (SABL)} to replace the conventional bounding box regression.
We extract side-aware features which focus on the content of boundaries for localization. A lightweight two-step \emph{bucketing scheme} is proposed to locate objects accurately based on the side-aware features.
We also introduce a rescoring mechanism to leverage the bucketing confidence to keep high-quality bounding boxes.
The proposed SABL exhibits consistent and significant performance gains on various object detection pipelines.

\noindent\textbf{Acknowledgement.}
This work is partially supported by the SenseTime Collaborative Grant on Large-scale Multi-modality Analysis (CUHK Agreement No. TS1610626 \& No. TS1712093), the General Research Fund (GRF) of Hong Kong (No. 14203518 \& No. 14205719), SenseTime-NTU Collaboration Project and NTU NAP.

%%%%%%%%%%%%%%%%%%%%%%%%%%%%%%%%%%%%%%%%%%%%%%%%%%%%%%%%%%%%%%%%%%%%%%%%%%%%%%%%%%%%%%%%%%%%%%%%%%%

%\clearpage
\bibliographystyle{splncs04}
\bibliography{egbib}
\clearpage
% !TEX root = ../2272.tex
\appendix
\section{Extensions of SABL in COCO Challenge 2019}
\label{sec:extension}
Here we demonstrate the whole system with bells and whistles in COCO Challenge 2019, which won the detection track of \emph{no external data}.
Applying SABL to Mask R-CNN with ResNet-50 \cite{He_2016} achieves 40.0\% $AP_{box}$ and 35.0\% $AP_{mask}$.
Then we adopt Hybrid Task Cascade (HTC) \cite{Chen_2019_CVPR} with the proposed CAFA (in Appendix ~\ref{sec:cafa}) in PAFPN \cite{liu2018_panet} and CARAFE \cite{Wang_2019_ICCV} in Mask Head.
It achieves 44.3\% $AP_{box}$ and 38.4\% $AP_{mask}$ compared with 42.1\% $AP_{box}$ and 37.3\% $AP_{mask}$ of HTC baseline.
Our overall system is trained without involving external instance-level annotated data during training. To be specific, it is trained on COCO2017 training split (instance segmentation and stuff annotations) as in \cite{Chen_2019_CVPR}. 
Here we also list other steps and additional modules we used to obtain the final performance. The step-by-step gains
brought by different components are illustrated in Table \ref{tab:step_by_step}.

\noindent\textbf{SyncBN}. We use Synchronized Batch Normalization \cite{liu2018_panet,Peng_2018} in the backbone and heads.

\noindent\textbf{SW}. We adopt Switchable Whitening (SW) \cite{pan2018switchable} in the backbone and FPN following the original paper.

\noindent\textbf{DCNv2}. We appply Deformable Convolution v2 \cite{Zhu_2019_CVPR} in the last three stage (from res3 to res5) of the backbone.

\noindent\textbf{Multi-scale Training}. We adopt multi-scale training. 
The scale of short edge is randomly sampled from $[400, 1400]$ per iteration and the scale of long edge is fixed as 1600.
The detectors are trained with 20 epoches and the learning rate is decreased by 0.1 after 16 and 19 epoches, respectively.

\noindent\textbf{SENet-154 with SW}. We tried different larger backbones. SENet-154 \cite{Hu_2018_SENet} 
with Switchable Whitening (SW) \cite{pan2018switchable} achieves the best single model performance. 

\noindent\textbf{Stronger Augmentation}. We adopt Instaboost \cite{fang2019instaboost} as the sixth policy of AutoAugment \cite{Cubuk_2019_CVPR}. 
Each policy has the same probability to be used for data augmentation during training procedure.
The detectors are trained with 48 epoches with such stronger augmentation, 
and the learning rate is decreased by 0.1 after 40 and 45 epoches, respectively.

\noindent\textbf{Multi-scale Testing}. We use 5 scales as well as horizontal flip at test time before ensemble. The testing scales are $(600, 900)$, $(800, 1200)$, $(1000, 1500)$, $(1200, 1800)$, $(1400, 2100)$.

\noindent\textbf{Ensemble}. We use ensemble of models based on five backbone networks.
We pretrain SENet-154 w/ SW and SE-ResNext-101 w/ SW on ImageNet-1K image classification dataset and use pretrained weights of ResNeXt-101 $32\times32$d, ResNeXt-101 $32\times16$d \cite{Xie_2017} and ResNeXt-101 $32\times8$d \cite{Xie_2017} provided by PyTorch
\footnote{\href{https://pytorch.org/hub/facebookresearch_WSL-Images_resnext/}{https://pytorch.org/hub/facebookresearch\_WSL-Images\_resnext/}}.

As shown in Table~\ref{tab:final results}, on COCO 2017 test-dev dataset, our method finally achieves 57.8\% $AP_{box}$, 51.3\% $AP_{mask}$ with multiple model ensemble and 56.0\% $AP_{box}$, 49.4\% $AP_{mask}$ with single model. Our result outperforms the 2018 COCO Winner Entry by 1.7\% $AP_{box}$ and 2.3\% $AP_{mask}$, respectively.

\begin{table}[t]
	\caption{
		Step by Step results of our method on COCO2017 \emph{val} dataset.
	}
	\label{tab:step_by_step}
	\begin{center}
		\begin{tabular}{c|c|cc}
			\hline
			Methods&scheduler&AP$_{box}$ &AP$_{mask}$ \\
			\hline
			Mask R-CNN & 1x & 37.3 & 34.2 \\
			+ SABL & 1x & 40.0 (+2.7) & 35.0 (+0.8) \\
			+ HTC & 1x & 42.9 (+2.9) & 37.4 (+2.4)  \\
			+ CAFA\&CARAFE & 1x & 44.3 (+1.4) & 38.4 (+1.0) \\
			+ SyncBN & 1x  & 45.8 (+1.5)& 39.9 (+1.5)  \\
			+ SW & 1x  &  46.1 (+0.3) & 40.0 (+0.1)  \\
			+ Backbone DCNv2 & 1x & 48.2 (+2.1) & 41.7 (+1.7) \\
			+ Mask Scoring & 1x & 48.3 (+0.1) & 42.4 (+0.7) \\
			+ MS-Training & 20e &  50.2 (+1.9) & 44.5 (+2.1) \\
			+ SE154-SW & 20e & 52.7 (+2.5)  & 46.1 (+1.6) \\
			+ AutoAug\&InstaBoost & 4x & 54.0 (+1.3) & 47.1 (+1.0)  \\
			+ Multi-Scale Testing & - & 55.3 (+1.3) & 48.4 (+1.3) \\
			+ Ensemble & - & \textbf{57.2} (+1.9) & \textbf{50.5} (+2.1) \\
			\hline
		\end{tabular}
	\end{center}
	\vspace{-10pt}
\end{table}

\begin{table}[t]
	\caption{
		Results with bells and whistles on COCO2017 \emph{test-dev} dataset.
	}
	\label{tab:final results}
	\begin{center}\small
		\begin{tabular}{c|cc}
			\hline
			Methods&AP$_{box}$ &AP$_{mask}$ \\
			\hline
			2018 Winners Single Model & 54.1 & 47.4 \\
			Ours Single Model & 56.0 & 49.4 \\
			\hline
			2018 Winners Ensemble~\cite{Chen_2019_CVPR} & 56.0 & 49.0 \\
			Ours & \textbf{57.8} & \textbf{51.3} \\
			\hline
		\end{tabular}
	\end{center}
\end{table}
\section{Content-Aware Feature Aggregation (CAFA)}
\label{sec:cafa}
Many studies~\cite{nas_fpn,kong2018deep,lin2017_fpn,liu2018_panet,pang2019libra} have investigated the architecture design for generating the feature pyramid
The approach for fusing low- and high-level features, however, remains much less explored.
A typical way for fusing features across different scales is by na\"{i}ve downsampling or upsampling followed by element-wise summation.
While this method is simple, it ignores the underlying content of each scale during the fusion process.

\begin{figure}[t]
	\centering
	\includegraphics[width=0.7\linewidth]{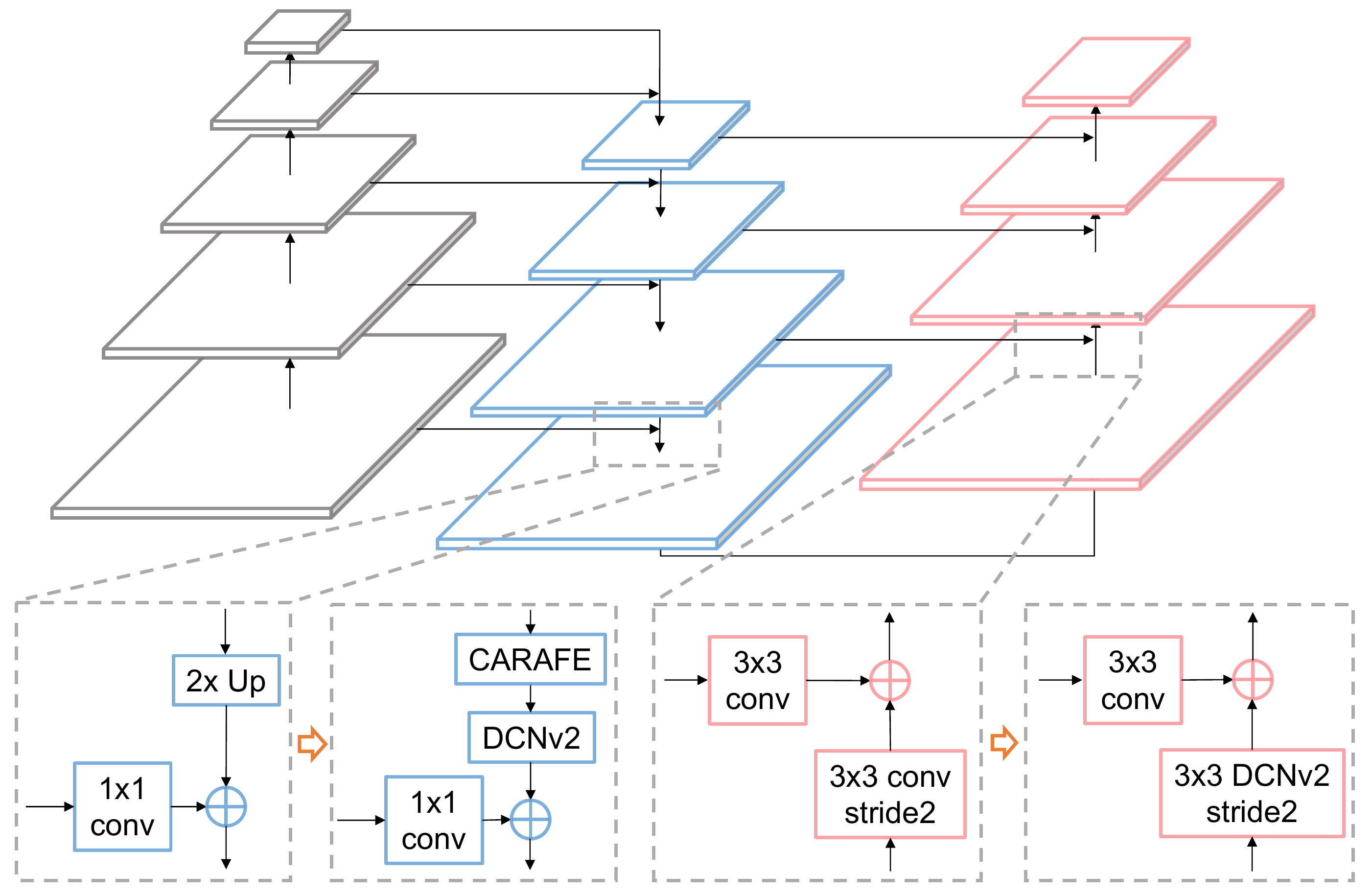}\\
	\caption{
		Modification by CAFA on PAFPN \cite{liu2018_panet}.
		We use CARAFE \cite{Wang_2019_ICCV} and DCNv2 \cite{Zhu_2019_CVPR} during upsampling
		and use DCNv2 \cite{Zhu_2019_CVPR} for downsampling.
	}
	\label{fig:uppafpn}
\end{figure}

Considering this issue, we propose the \textbf{Content-Aware Feature Aggregation} (CAFA) module that facilitates effective fusion and aggregation of multi-scale features in a feature pyramid.
To encourage content-aware upsampling, we develop CAFA based on CARAFE~\cite{Wang_2019_ICCV}, a generic operator which is proposed for replacing the conventional upsampling operator.
Different from CARAFE, we perform Deformable Convolution v2~\cite{Zhu_2019_CVPR} after CARAFE while before summing up the upsampled feature maps with the lateral feature maps (see Figure~\ref{fig:uppafpn}).
\begin{table}[ht]
	\caption{
		Effectiveness of CAFA combined with FPN \cite{lin2017_fpn} and PAFPN \cite{liu2018_panet} in Mask R-CNN \cite{mask_rcnn}.
	}
	\label{tab:upfpn}
	\begin{center}
		\begin{tabular}{c|l|cc}
			\hline
			Method&Modification&box AP&mask AP \\
			\hline \multirow{2}{*}{FPN}
			&Baseline           &37.3  &34.2 \\
			&+ CAFA &38.9  &35.3 \\
			\hline \multirow{2}{*}{PAFPN}
			&Baseline           &37.7  &34.3 \\
			&+ CAFA &40.0  &36.2 \\
			\hline
		\end{tabular}
	\end{center}
\end{table}
Further, the conventional convolution layers are replaced by Deformable Convolution v2~\cite{Zhu_2019_CVPR} for downsampling.
Thanks to this design, CAFA enjoys a larger receptive field and gains improved performance in adapting to instance-specific contents.  

As shown in Table~\ref{tab:upfpn}, we study the effectiveness of CAFA combined with 
FPN \cite{lin2017_fpn} and PAFPN \cite{liu2018_panet} in Mask R-CNN \cite{mask_rcnn}.
CAFA brings 1.6\% $AP_{box}$, 1.1\% $AP_{mask}$ gains on FPN, and 2.3\% $AP_{box}$, 1.9\% $AP_{mask}$ gains on PAFPN, respectively.

We further evaluate CAFA via comparing it with NAS-FPN on RetinaNet and it achieves compatible results (39.2\% $AP_{box}$ v.s. 39.5\% $AP_{box}$ on COCO2017 \emph{val} dataset at $640\times640$ scale).
While NAS-FPN uses 7 pyramid networks, our CAFA with PAFPN only uses 2 pyramid networks 
with much simpler pathways (one top-down and one bottom-up) among pyramidal features.

\section{Visual Results Comparison}
\begin{figure}[]
	\centering 
	\includegraphics[width=\linewidth]{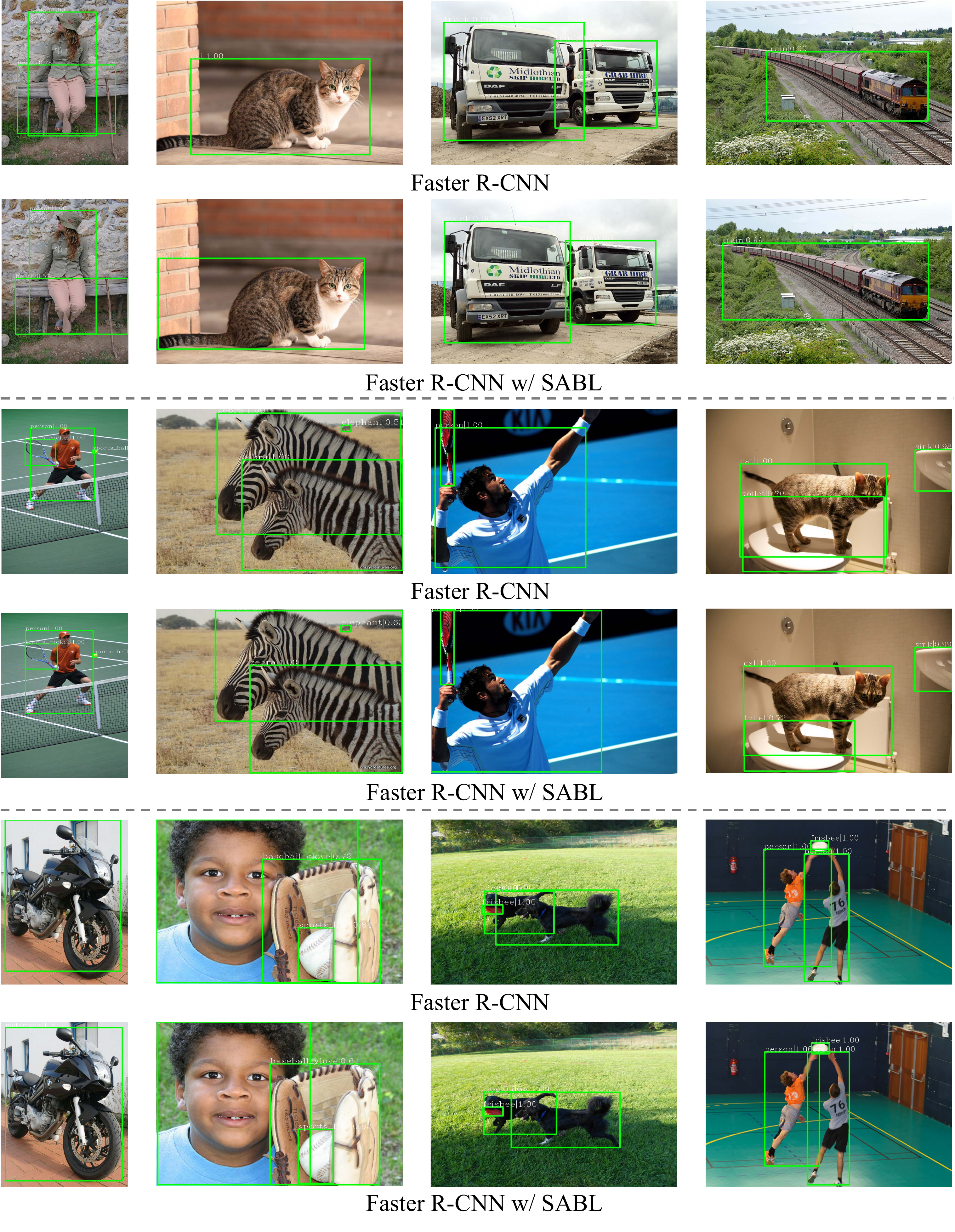}
	\caption{
		\small{Comparison of object detection results between Faster R-CNN baseline and Faster R-CNN w/ SABL on COCO 2017 val. ResNet-101 w/ FPN backbone and 1x training schedule are adopted in both methods. Faster R-CNN w/ SABL shows more precise localization results than the baseline.}
	}
	\label{fig:detection}
\end{figure}

As illustrated in Figure~\ref{fig:detection}, we provide some object detection results comparison between Faster R-CNN~\cite{ren2015faster} baseline and Faster R-CNN w/ SABL on COCO 2017~\cite{lin2014coco} val. ResNet-101 w/ FPN backbone and 1x training scheduler are adopted in both methods. Faster R-CNN w/ SABL shows more precise localization results than the baseline.
\clearpage

\end{document}